\documentclass{article}

\usepackage[final]{corl_2020} 
\usepackage{graphicx}
\usepackage{wrapfig}
\graphicspath{{./figs/}}

\title{Bayes-Adaptive Deep Model-Based Policy Optimisation}

\usepackage{multirow}
\usepackage{booktabs}

 \usepackage{amsmath}
  \usepackage{amssymb}
  \usepackage{amsfonts}

\newcommand{\cS}{{\mathcal S}}
\newcommand{\cA}{{\mathcal A}}

\newcommand{\cD}{{\mathcal D}}

\newcommand{\cT}{{\mathcal T}}
\newcommand{\cR}{{\mathcal R}}

\newcommand{\cH}{{\mathcal H}}

\newcommand{\cN}{{\mathcal N}}
\newcommand{\cL}{{\mathcal L}}

\newcommand{\ve}[1]{\mathbf{#1}}

\usepackage{algorithm}
\usepackage{algpseudocode}
\algrenewcommand{\algorithmicrequire}{\textbf{Input:~~}}
\algrenewcommand{\algorithmicensure}{\textbf{Output:}}
\algrenewcommand{\algorithmiccomment}[1]{\qquad\hfill~\hspace*{-5ex}\textit{// #1}}

\DeclareMathOperator*{\argmax}{arg\,max}

\newcommand{\KLD}[2]{D_{\mathrm{KL}} \left( \left. \left. #1 \right|\right| #2 \right) }

\author{
  Tai Hoang\\
  Technical University of Munich\\
  Munich, Germany\\
  \texttt{t.hoang@tum.de} \\
   \And
   Ngo Anh Vien \\
   Bosch Center for Artificial Intelligence \\
   Renningen, Germany \\
   \texttt{anhvien.ngo@bosch.com} \\
}

\begin{document}
\maketitle


\begin{abstract}
We introduce a Bayesian (deep) model-based reinforcement learning method (RoMBRL) that can capture model uncertainty to achieve sample-efficient policy optimisation. We propose to formulate the model-based policy optimisation problem as a Bayes-adaptive Markov decision process (BAMDP). RoMBRL maintains model uncertainty via belief distributions through a deep Bayesian neural network whose samples are generated via stochastic gradient Hamiltonian Monte Carlo. Uncertainty is propagated through simulations controlled by sampled models and history-based policies. As beliefs are encoded in visited histories, we propose a history-based policy network that can be end-to-end trained to generalise across history space and will be trained using recurrent Trust-Region Policy Optimisation. We show that RoMBRL outperforms existing approaches on many challenging control benchmark tasks in terms of sample complexity and task performance. The source code of this paper is also publicly available on \url{https://github.com/thobotics/RoMBRL}.
\end{abstract}

\keywords{Model-based RL, policy optimization, Bayes-adaptive MDP} 


\section{Introduction}
The family of reinforcement learning (RL) algorithms consists of two major approaches: \emph{model-free} and \emph{model-based} \cite{sutton2018reinforcement}. The model-free approach does not attempt to estimate a model of the environment and is known to require less tuning. However, they are known to suffer from a high variance problem. On the contrary, the model-based approach tends to have a lower sample complexity than the model-free approach. Hence it has a wider application to practical tasks, e.g. robotics \cite{DeisenrothR11,LevineFDA16}. Model-based approaches alternate between model learning and policy optimisation. Model learning is based on samples generated from real interactions with the environment. The policy is optimised using fictitious data generated from interactions with the learnt model. On large domains, there is deep model-based learning \cite{KurutachCDTA18,ChuaCML18}. However, the online policy optimisation step might be based on biased samples that are imagined from a very inaccurate model, as discussed by Atkeson and Santamaria \cite{atkeson1997comparison}. It is well-known that both the standard model-based and model-free approaches suffer from a poor trade-off between exploration and exploitation. The main reason is at their weak reasoning ability over the uncertainty about the unknown dynamics of the environment \cite{ghavamzadeh2015bayesian}.

In this paper, we will formulate Bayesian model-based RL (MBRL) as a principled Bayes-adaptive Markov decision process (BAMDP). Essentially, BAMDP transforms an RL problem into a belief planning problem, i.e. planning in partially observable MDP (POMDP) where beliefs are modeled as probability distributions over the environment's states. Common simulation-based belief planning solvers alternate between three major steps: i) execute the policy in the environment to collect real data in order to update the belief model; ii) simulate the sampled dynamics from the belief distribution; and iii) update the policy using fictitious data received from the simulations. We use a Bayesian neural network (BNN) to maintain an uncertainty model (\emph{belief} estimation) based on the data collected from interactions with the environment. Many existing methods are able to provide uncertainty estimation and still enjoy the scalability of training with deep neural networks, including variational inference methods \cite{Graves11,liu2016stein}, expectation propagation \cite{Hernandez-Lobato15b}, approximate inference via dropout \cite{gal2016dropout}, and other stochastic gradient-based MCMC methods \cite{balan2015bayesian,ChenFG14}. Among these methods, stochastic gradient Hamiltonian Monte-Carlo (SGHMC) \cite{ChenFG14} is considered to be a scalable method with well-calibrated uncertainty which is a desirable feature for the contexts requiring high-quality uncertainty estimation as required by planning under uncertainty. We will use SGHMC to sample from a posterior distribution over the environment dynamics.

We posit to use \emph{root-sampling} \cite{GuezHSD14} via SGHMC to avoid computational belief updates. We then use the simulation-based approach to build a \emph{search tree of beliefs} in which each node represents a history of observations and actions. Uncertainty propagation is managed by the sampled dynamics (\emph{fictitious simulation}) along the tree using a recurrent policy that is defined as a mapping from a history of observation-action pairs to a next action, instead of being from a true state to an action as seen in MDP domains. The recurrent policy enables the generalisation of action selections across multiple histories in the belief space. Fictitious simulations are then fed to the training of the recurrent policy using Long Short-Term Memory (LSTM)-based Trust-Region Policy Optimisation (TRPO). Putting them all together, we propose robust Bayesian model-based reinforcement learning (RoMBRL), a principled Bayesian deep model-based RL framework.

We demonstrate RoMBRL on a set of Mujoco control tasks \cite{1606.01540}. Experiment results show that the proposed RoMBRL offers an effective reasoning ability about uncertainty over the environment dynamics. When compared to other baseline methods, RoMBRL obtains a better sample complexity and a higher quality policy in terms of task performance.

\section{Related Work}
Earlier MBRL methods are mostly based on simple linear model parameterisation \cite{schneider1997exploiting,atkeson1997comparison,AbbeelQN06,LevineA14,LevineFDA16}. Therefore they might have limited representation power and do not scale to complicated dynamics models and large data-sets. Besides, these approaches have a limited reasoning ability about the uncertainty over the environment dynamics, which leads to inefficient sample complexity. Alternative methods use an ensemble model to estimate uncertainty \cite{RajeswaranGRL17,ChuaCML18,KurutachCDTA18,ShyamJG19} and handle policy optimisation correspondingly. Although these approaches have shown many promising results on many complex control tasks, they do not provide high-quality uncertainty estimation hence sometimes result in poor performance and have high sample complexity. 

Many previous works have used Bayesian inference for model-based RL, e.g. look-up tabular approaches \cite{PoupartVHR06,WangWHL12,VienE12,GuezSD13,VienT15} (see \cite{ghavamzadeh2015bayesian} for more details). A common formulation to Bayesian MBRL is to transform the Bayesian MBRL problem into a planning problem under the partially observable Markov decision process (POMDP) framework. The belief update becomes tractable only when the problem is small and discrete. A more effective method is using Gaussian processes \cite{DeisenrothR11} that can be very data-efficient, however not able to capture non-smooth dynamics. On domains with non-smooth dynamics, Gaussian mixture models (GMMs) can be an alternative \cite{LevineA14}. However both of them are too computationally expensive on large data-sets. Recent works use variational inference \cite{DepewegHDU17,IglZLWW18,okada2020variational} and dropout \cite{GalHK17} as Bayesian methods for estimating model uncertainty. However variational inference or dropout is known to have suboptimal uncertainty estimates \cite{SpringenbergKFH16}. 

A very closely related idea to our work is recently proposed by Lee et. al. \cite{Gilwoo19} in which Bayesian MBRL is also formulated as a POMDP. Then a standard policy optimisation method is used to compute a policy which can generalise action selection across the belief space. However, their approach must hand-engineer a dynamics representation using \emph{domain knowledge} and \emph{belief discretisation}, hence the belief could only model a low-dimensional distribution over a finite set of environment parameters. In contrary, our approach directly maintains a belief over the environment dynamics using a general Bayesian neural network.

\section{Background}
\subsection{Model-Based Reinforcement Learning}
\label{sec_mbrl}
Markov Decision Process (MDP) is a mathematically principled framework that can model a sequential decision making process of an agent that interacts with an environment. A MDP is defined as a five-tuple $M =\{\cS,\cA,\cT,\cR, \gamma\}$. Given a MDP model, an agent's goal is to find an optimal policy defined as a stochastic mapping $\pi:\cS \times \cA \mapsto [0,1]$ so that a cumulative discounted reward as a performance measure, $
  J(\pi) = \mathbb{E}_{\pi,\cT} \Big[\sum_t \gamma^t r_t\Big]$ is maximised w.r.t the stochasticity of the environment's dynamics and the policy.

If the environment dynamics $\cT$ is unknown to the agent, MBRL can be used to find an optimal policy. MBRL estimates $\cT$ by a parametric function $s'=f_\theta(s,a)$ of parameter $\theta$.\footnote{We assume a deterministic dynamics} For example, given a training data-set $\cD=\{s_t,a_t,s_{t+1}\}_{t=0}^{|\cD|}$, an objective of model learning can be the mean square error of predictions: 
$  L_{\text{model}} (\theta) = \frac{1}{|\cD|} \sum_t \|s_{t+1} - f_\theta(s_t,a_t) \|_2^2 .
 $
The learnt model can be used as a simulator to generate fictitious data to train a policy function $\pi_w(s,a)$ of parameters $w$. For example, \cite{KurutachCDTA18} propose model-ensemble trust-region policy optimisation (ME-TRPO, see Algorithm \ref{me-trpo}). ME-TRPO fits a set of $k$ dynamics functions $f_\theta=\{f_{\theta_1},f_{\theta_2},\ldots,f_{\theta_k}\}$ on the same data-set $\cD$, each $f_{\theta_i}$ is modelled with a feed-forward network. Model learning is a standard supervised learning problem, ME-TRPO uses a standard Adam optimiser \cite{KingmaB14} and other training techniques to avoid over-fitting. Using the re-parametrization trick \cite{HeessWSLET15}, a Gaussian policy can be re-written to sample $m$-dimensional actions from a parameterized distribution $\mu_w(s) +\sigma_w(s)^\top \zeta$, where $\zeta \sim \cN(0,I_m)$, and $I_m$ is an identity matrix of dimensionality $m$. Another feed-forward neural network is used to represent $\mu_w$ while $\sigma_w$ can be additional outputs of the mean estimator network or manually tuned. ME-TRPO proposes to use Trust-Region Policy optimisation (TRPO) \cite{SchulmanLAJM15} to train the policy network using fictitious data generated from the trained dynamics model $f_\theta$. The gradient is computed using likelihood-ratio methods as $
  \nabla_w J(\pi) \approx \frac{1}{N}\sum_{i=1}^N \left[\nabla_w p(\xi_i;\theta) R(\xi_i) \right] 
$, where $p(\xi_i;\theta)$ is a trajectory probability distribution, $R(\xi_i)$ is the discounted return of trajectory $\xi_i$, and $\{\xi_i\}_{i=1}^N$ is the fictitious trajectory set generated using policy $\pi_w(s,a)$ on the learnt model $f_\theta$. A vanilla model-based  deep reinforcement learning is a special case of ME-TRPO when the number of models $k$ in the ensemble is set to one.


  \begin{algorithm}

  \caption{Model-Ensemble Trust-Region Policy Optimisation (ME-TRPO)}
  \label{me-trpo}
  \begin{algorithmic}[1]
    \State Initialise policy $\pi_w$ and all dynamics $f_\theta=\{f_{\theta_i}\}_{i=1}^k$
    \State Initialise a data-set $\cD=\emptyset$
    \Repeat
    \State Rollout trajectories $\{s_t,a_t,s_{t+1}\}_{t=0}^T$ using $\pi_w$
    \State Update: $\cD = \cD \cup \{s_t,a_t,s_{t+1}\}_{t=0}^T$
    \State Train all dynamics model in $f_\theta$ using $\cD$ 
    \Repeat
    \State Collects $N$ trajectories $\{\xi_i\}_{i=1}^N$ from dynamics $f_\theta=\{f_{\theta_i}\}_{i=1}^k$ using $\pi_w$
    \State Update policy $\pi_w$ on fictitious data $\{\xi_i\}_{i=1}^N$ 
    \Until policy $\pi_w$ fail the validation
    \Until $\pi_w$ is optimal    
\end{algorithmic}
  \end{algorithm}


\subsection{Bayes-Adaptive Markov Decision Process}
\label{bayes}
Bayesian model-based RL in the underlying Markov decision process $M$ can be formulated as a Bayes-adaptive MDP (BAMDP) which is formally defined as a five-tuple $M^+=\{\cS^+,\cA,\cT^+,\cR^+, \gamma\}$ as suggested by \cite{Duffphdthesis,GuezSD13}, where $S^+=\cH \times \cS$ ($\cH$ is the space of all possible histories of states, actions, and rewards; $\cS$ is the state space of the underlying MDP $M$). It can be shown that the optimal policy on $M^+$, e.g. using dynamic programming, is a (Bayesian) optimum policy for exploration-exploitation trade-off on an unknown MDP $M$ \cite{PoupartVHR06,GuezSD13}. In particular, we denote $p(f)$ as a prior distribution over the unknown dynamics of $M$, and compute a belief as a posterior distribution $b_t(f) = p(f|h_t)$ conditional on history $h_t$ up to time $t$. The belief $b_t(f)$ represents the uncertainty about the dynamics of the model $M$. Based on belief modelling, we can compute the dynamics $\cT^+$ and reward functions $\cR^+$ of the Bayes-adaptive MDP model $M^+$ as
\begin{equation}
\begin{aligned}
\cT^+\left( (h,s),a,(has',s')\right)= p\left((has',s')|(h,s),a\right) = \int_f p_f(s'|s,a)p(f|h ) df; \quad R^+ (h,s,a) = \cR(s,a)
\end{aligned}
 \label{pomdp}
 \end{equation}
where $p_f$ is a Dirac function if a deterministic dynamics model is assumed like in ME-TRPO. 

Approximating policies in $M^+$ can be performed via tabular look-ups \cite{GuezSD13} or policy graphs \cite{WangWHL12} for discrete MDP, or value function approximations \cite{GuezHSD14}. For an example, Guez et. al. \cite{GuezHSD14} propose the Bayes-adaptive simulation-based planning algorithm (BAMCP). BAMCP handles the search via \emph{root sampling}, where models are sampled once at the root node, instead of being sampled at every tree node to avoid computational belief updates. BAMCP uses a function approximator $Q_w(h,s,a)$ of particular history $h$, state $s$ and action $a$. Using function approximation can help generalise value estimation over infinite-dimensional and continuous belief space. In each online planning step at time $t$, BAMCP samples $k$ dynamics models from the posterior $p(f|h_t)$ which are used to simulate fictitious trajectories in order to compute value function estimates. Within each simulation, BAMCP proposes to use an incremental or online algorithm to update $w$, e.g. Q-learning or gradient temporal difference.


Recently, there is a similar effort to scale BAMCP through the use of deep neural networks. \cite{Gilwoo19} propose Bayesian policy optimisation (BPO) whose very significant extension is to replace the value function approximation $Q_w(h,s,a)$ by a policy network. Based on fictitious simulations, the policy network can be trained using a standard policy optimisation method, e.g. TRPO. Though using gradient-descent could only learn a locally optimal policy, BPO has shown promising results on many continuous tasks that could not be solved previously by BAMCP. However, both BAMCP and BPO rely on hand-designed belief representation (e.g. a small finite set of unknown environment parameters like the mass or length of a robot joint etc.), hence very domain-specific and have limited applicability.


\section{Robust Bayesian Model-Based Reinforcement Learning}
We propose a general-purpose Bayesian MBRL, a robust Bayesian model-based reinforcement learning (RoMBRL), which combines \emph{i}) forward-search via \emph{root-sampling}, \emph{ii}) model learning via a Bayesian neural network, and \emph{iii}) recurrent policy optimisation via LSTM-based TRPO.

\subsection{Algorithm}
Similar to BAMCP, RoMBRL's policy learning is handled as Bayes-adaptive planning by building a \emph{search tree} of belief states. Each belief node represents a visited history $h_t$ and an \emph{optimal action} $a_t$. Similar to BPO \cite{Gilwoo19}, RoMBRL represents the optimal actions directly at each belief node. RoMBRL uses a recurrent policy $\pi_w: \cH \times \cS\times \cA \mapsto[0,1]$, instead of a value function approximation used in BAMCP. The distribution written as $\pi_w(\cdot|h_t,s_{t})$ represents a policy of particular history $h_t$ and state $s_{t}$, where $w$ are weights of a deep recurrent neural network. $\pi_w(a_{t}|h_t,s_{t})$ can generalise action selection across beliefs that are encoded in histories of visited states and actions $h_t=\{s_0,a_0,\ldots, a_{t-1}\}$. Instead of using computationally expensive full planning like MCTS \cite{GuezSD13} or point-based value iteration methods \cite{PoupartVHR06}, we resort to a gradient-based approximation method to estimate policies, i.e. recurrent or history-based policy optimisation \cite{wierstra2010recurrent}.

A concise summary of RoMBRL is described in Algorithm~\ref{bapo}. RoMBRL handles planning by sampling from posterior $p(f_\theta|\cD)$ via root-sampling. For each sampled dynamics model $f_\theta$, RoMBRL simulates a trajectory $\{s_0,a_0,\ldots,s_T\}$. Actions are selected according to a policy $\pi_w(a_t|h_t,s_t)$. Fictitious state transitions are generated from the sampled dynamics model $f_\theta$. Note that we are using root-sampling to avoid sampling directly from the belief dynamics $\cT^+\left( (h,s),a,(has',s')\right)$ that requires a computationally expensive belief update as shown in Eq.~\ref{pomdp}. Guez et. al. \cite{GuezHSD14} have shown that the rollout distributions resulting from either root-sampling (sampling once at the root) or the one with belief updates (sampling at every tree nodes after its corresponding belief gets updated) are the same. RoMBRL waits for $k$ complete rollouts to update its policy $\pi_w$ using a standard policy optimisation framework, i.e. through TRPO with a recurrent policy network.

\begin{algorithm}[t]
  \caption{Robust Bayesian model-based reinforcement learning (RoMBRL)}
  \label{bapo}
  \begin{algorithmic}[1]
    \State Initialise policy $\pi_w$ and prior distribution $p(f_\theta)$
    \State Initialise data-set $\cD=\emptyset$, iteration $l=0$
    \Repeat
    \State Collect real trajectories $\{s_t,a_t,s_{t+1}\}_{t=0}^T$ using $\pi_w$
    \State Update: $\cD_{l+1} = \cD_l \cup \{s_t,a_t,s_{t+1}\}_{t=0}^T$ \Comment{{\color{green}Model learning}: on real data}
    \For{\texttt{$i=1:k$}}
    \State Sample a dynamic: $f_{\theta_i} \sim p(f_\theta |\cD)$ \Comment{{\color{red}Sampling from the model posterior}. See \ref{sampling}}
    \State Store $H_i=\text{SIMULATE}(\emptyset,s_0,\pi_w,f_{\theta_i})$ 
    \EndFor    
    \State Update $\pi_w$ on fictitious data $\{H_i\}_{i=1}^k$  \Comment{{\color{blue}Policy learning}: on fictitious data. See \ref{seerpo}}
    \State $l=l+1$
    \Until $\pi_w$ is optimal
    \\
    \Procedure{Simulate}{$h,s_0,\pi,f$} \Comment{Simulate on the sampled dynamics model}
    \State $t=0$; $H=\emptyset$; $s=s_0$
    \While {$t<T$}
    \State Select an action $a\sim \pi(\cdot|h,s)$
    \State Execute $a$: $s'= f(s,a),\ r = \cR(s,a)$ \Comment{A fictitious transition}
    \State $H = H \cup (h,a,has')$ \Comment{History collection} 
    \State $t=t+1; s=s'; h= has'$
    \EndWhile
    \State \textbf{return} $H$
     \EndProcedure
  \end{algorithmic}
\end{algorithm}


\paragraph{Connection to BAMCP}Though inspired by the BAMCP framework, our proposed RoMBRL is different from BAMCP in many aspects. BAMCP assumes a discrete action space therefore a value function $Q_w(h,s,a)$ can be used to select actions. Besides, BAMCP uses manually-designed features; therefore it is non-trivial to train with a large data-set. Moreover, the increment update methods used by BAMCP, e.g. SARSA or Q-learning, can be costly if being scaled to large parametric models, e.g. deep neural networks. RoMBRL overcomes these challenges by using a direct policy representation $\pi_w(a|h,s)$ and stochastic policy gradient approaches, i.e. TRPO. Different from BPO and BAMCP, which assumes a domain-specific belief representation, RoMBRL uses a general-purpose Bayesian neural network to maintain beliefs, which is more \emph{task-agnostic}. 

\paragraph{Connection to Ensemble Methods}
One simple explanation to ME-TRPO \cite{KurutachCDTA18} and PETS \cite{ChuaCML18} in a Bayesian way is to treat the model ensemble $f_\theta=\{f_{\theta_i}\}_{i=1}^k$ as particle filtering used to represent the posterior $p(f_\theta|\cD)$, $p(s'|s,a) \approx \frac{1}{k}\sum_{i=1}^k p(s'|s,a;f_{\theta_i})$,
where we assume $p(s'|s,a;f_{\theta_i})=\delta_{s'}(f_{\theta_i}(s,a))$ \footnote{$\delta$ is an identity function} in the case of deterministic function $f_\theta$. These particles are with uniform weights which are \emph{not updated}. Therefore they can not approximate \emph{accurately} the posterior distribution over the environment dynamics. While PETS is able to do planning under uncertainty via model predictive control, ME-TRPO is not designed to do so. As PETS uses model predictive control (MPC) and cross entropy method (CEM) as action selection and our paper is mainly focused on using policy optimization (TRPO) to update the policy, therefore we decide to treat only ME-TRPO as the main baseline.

\subsubsection{Model Learning via Stochastic Gradient Hamiltonian Monte-Carlo}
\label{sampling}
We use SGHMC proposed by Springenberg et. al. \cite{SpringenbergKFH16} to sample from the posterior distribution $p(\theta|\cD)$ of parameters $\theta$ given a data-set $\cD=\{s_t,a_t,s_{t+1}\}_t$ as seen in Step 7 in Algorithm \ref{bapo}. 
We define a probabilistic model for belief functions over dynamics $s'=f_\theta(s,a)$ as
\begin{align}
  p(f(s,a)|s,a,\theta) = \cN(f(s,a;\theta_\mu);\theta_{\sigma^2}),
  \label{bmodel}
  \end{align}
where we denote $\theta=[\theta_\mu , \theta_{\sigma^2}]$ as model parameters, and we assume a homoscedastic noise with zero mean and variance $\theta_{\sigma^2}$. A sample $f(s,a;\theta_\mu)$ from the above distribution results in one parametric dynamics model with parameters $\theta_\mu$.

Samples are generated from a joint distribution of $\theta$ and $r$ as, $
  p(\theta,r|\cD) \propto \exp \left(  -U(\theta) -\frac{1}{2} r^\top \ve{M}^{-1} r\right )$, 
where $r$ is an auxiliary variable, $U(\theta) = - \log p(\cD|\theta) $ is the \emph{potential energy} function which is set to  the negative log-likelihood. HMC samples $\theta$ by simulating a physical system described by Hamilton's equations $H(\theta,r) =  -U(\theta) -\frac{1}{2} r^\top \ve{M}^-{1} r$, where $r$ is called the momentum of the system, and $\ve{M}$ is the mass matrix. The simulation requires computing the costly gradient $\nabla U(\theta)$ on a complete data-set $\cD$. There is an idea \cite{ChenFG14} proposing to use a noisy estimate based on a \emph{minibatch}, $\nabla \tilde U{\theta} \approx U(\theta) + \cN(0,2(\ve{C}-\ve{\hat B})\epsilon)$, then use $\epsilon$-discretization to receive a discrete-time system as
\begin{equation}
\begin{aligned}
  \Delta \theta = \epsilon \ve{M}^{-1} r\quad \Delta r = -\epsilon \nabla \tilde U(\theta) - \epsilon \ve{C} \ve{M}^{-1} r + \cN(0,2(\ve{C}-\ve{\hat B})\epsilon),
\end{aligned}
  \label{sghmc}
  \end{equation}
where $\ve{C}$ is the friction term, $\ve{\tilde B}$ is an estimate of the diffusion matrix resulting from the gradient noise, and $\epsilon$ is a step size. HMC with minibatch is called Stochastic Gradient HMC (SGHMC), which is regarded as an efficient Bayesian learning method on deep neural network and large data-sets. To improve the robustness of model learning, the hyper-parameters $\ve{M}, \ve{c}, \ve{\tilde B}$, and $\epsilon$ are adaptively tuned during the burn-in process (i.e. mixing time in Markov chain) in which initial samples are discarded when the sampling has not yet converged to samples of the true posterior (to a stationary distribution).

\subsubsection{Recurrent Policy Optimisation}
\label{seerpo}
We now discuss Step 10 in Algorithm~\ref{bapo} on how policy optimisation for $\pi_w$ is handled.
We extend TRPO to find optimal policies in belief space, which has also been previously tried by \cite{abs-1810-07900}. Essentially, we replace the state space in TRPO for MDP by the augmented belief state space as defined in the previous section. As a result, TRPO updates recurrent policies $\pi_w$ of parameter $w$ at iteration $m$ with a trust region constraint as follows:
\begin{align*}
  w_{m+1} = \argmax_{w} \cL(w_m,w) \quad \quad  \text{s.t. } \quad \bar D_\mathrm{KL}(w||w_m) \le \delta ,
  \end{align*}
where both the \emph{surrogate advantage} $\cL(w_m,w)$ and the \emph{average KL-divergence} $\bar D_\mathrm{KL}(w||w_m)$ are based on histories,
\begin{align*}
  &\cL(w_m,w) = \mathbb{E}_{h,s,a}\left[\frac{\pi_w(a|h,s)}{\pi_{w_m}(a|h,s)} A_{\pi_{w_m}}(h,s,a) \right]
  \\
  &\bar D_\mathrm{KL}(w||w_m) =\mathbb{E}_{h,s}\left [ \KLD{\pi_w(\cdot|h,s)} {\pi_{w_m}(\cdot|h,s}\right ] .
\end{align*}
where $A_{\pi_{w_m}}(h,s,a)=\big( Q_{\pi_{w_m}}(h,s,a) - V_{\pi_{w_m}}(h,s)\big)$ is an advantage value function over $\pi_{w_m}$ of particular history $h$, state $a$, and action $a$. In our implementation, we simply use recurrent neural network layers in TRPO to represent $\pi_{w}$. 

\subsection{Online Bayesian Model Learning}
\label{onlinesampling}
Each iteration $l$ of model learning in Algorithm \ref{bapo} requires re-sampling from an updated data-set $\cD_{l+1} = \cD_{\text{l}} \cup \cD_{\text{new}}$, where $\cD_{\text{new}}=\{s_t,a_t,s_{t+1}\}_{t=0}^T$. A standard application of SGHMC as described in detail in \ref{sampling} just does the sampling on $p(\cdot|\cD)$ by starting from scratch (for each iteration $l$) with freshly initialised hyper-parameters $\ve{M}, \ve{c}, \ve{\tilde B}, \epsilon$ and the potential energy $U(\theta) = -\log p(\cD|\theta)$, in which $k$ dynamics $f_{\theta_i}$ are sampled.

A Bayesian approach offers online updates naturally using Bayes's rule \cite{opper1998bayesian}: $p(\theta|\cD_{\text{new}},\cD_{\text{l}}) \propto p(\cD_{\text{new}}|\theta) p(\theta|\cD_{\text{l}})$. Note that we can also sample from $p(\theta|\cD_{l+1})$ by initialising $\ve{M}, \ve{c}, \ve{\tilde B}, \epsilon$ to their final values at iteration $l$. However sampling this way even yields a worse performance than starting SGHMC from scratch. In our case, we divide the posterior $p(\cD_{l+1}|\theta)$ over the whole data into two parts: $p(\cD_\text{new}|\theta)$ and a cost that the new parameter must also fit the old data $\cD_\text{l}$. At iteration $l+1$, we first exploit the fitting distribution with parameters $\theta_l$ in previous iteration $l$, $p(\theta|\cD_\text{l})$. In particular, we use approximation by assuming that $p(\theta|\cD_l)$ is a normal prior with covariance $\Sigma$ where $\Sigma^{-1} \propto F(\theta_l)$, where $F(\theta_l)$ is the Fisher information matrix for $p(\theta_l|\cD_l)$. Then, we receive 
$
\log p(\theta|\cD_l) \propto \frac{1}{2}(\theta-\theta_l)^\top F(\theta_l)(\theta-\theta_l) + const
$, where $const$ consists of terms independent of $\theta$. This approximate log-likelihood is exactly the second-order approximation to the KL divergence $D_{\text{KL}} \left (p(\cD_l|\theta_l) \| p(\cD_l|\theta) \right)$ \cite{abs-1812-02256}.


We now propose an online Bayesian model learning version that uses SGHMC with the potential energy as $p(\theta|\cD_{l+1}) \propto p(\cD_\text{new}|\theta)p(\theta|\cD_{\text{l}})\approx \exp(-U(\theta))$ where 
\begin{align}
\label{eq:online_loss}
    U(\theta) = - \log p(\mathcal{D}_{\text{new}} | \theta)  + D_{\text{KL}} \left (p(\cD_l|\theta_l) \| p(\cD_l|\theta) \right )
\end{align}
As a result, $U(\cdot)$ can be approximated using samples $\theta_i \sim p(\cdot|D_{\mathrm{l}})$ (samples received at iteration $l$ using SGHMC) as
\begin{equation}
\begin{aligned}
\label{eq:online_loss2}
    U(\theta) \approx - \log p(\mathcal{D}_{\text{new}} | \theta) + \sum_{i=1}^k p(\mathcal{D}_{\text{l}} | \theta_i) \big( \log p(\mathcal{D}_{\text{l}} | \theta_i) - \log p(\mathcal{D}_{\text{l}} | \theta) \big )
\end{aligned}
\end{equation}
Note that without the KL-divergence part, SGHMC would sample only from a local model fitted to the most recent data ${D}_{\text{new}}$. Being inspired by Gal et. al. \cite{gal2016improving}, the potential energy in Eq.~\ref{eq:online_loss2} aims to weight more critical on the recent data $\cD_{\text{new}}$ rather than the entire collected ones. The local model fitted on data ${D}_{\text{new}}$ represents the true higher reward state as the policy gets improved. However, training the dynamics on such a small set of data might lead the model to the well-known catastrophic phenomenal, the \emph{biasing} problem. Thus, to avoid this problem as well as maintaining a more meaningful dynamic, we add one more term in the potential energy, $D_{\text{KL}} \left (\theta_l \| \theta \right )$, to force the current dynamics not stay too far away from the recently learned dynamics, encoded in $\cD_\mathrm{l}$. We observe that using the biased energy in Eq.\ref{eq:online_loss2} helps the burn-in process and improve the sample quality significantly.


\section{Experiments}
We evaluate RoMBRL on different scenarios of MBRL. Our experiments set to answer the following questions: i) How uncertainty estimation via the principled Bayesian neural network offer in comparison to the ensemble approach when learning a dynamics model? ii) Can a full Bayes-adaptive deep model-based RL method reduce sample complexity? iii) Can Bayesian model-based RL be promising for transfer learning w.r.t changing environment dynamics?

The third question comes from one of the advantages of a model-based approach in which a learnt dynamics model can be transferred to another task with a different reward function. Experiment setting details are put in appendix. All results are averaged over three runs with different random seeds.

\paragraph{Environment:}We used three different types of domains: i) A toy domain: We aim to analyse the behaviour of our Bayesian method on learning a simple transition distribution where only little data are observed. ii) Four Mujoco locomotion tasks \cite{1606.01540}: \emph{Swimmer}, \emph{Hopper}, \emph{Half Cheetah} and \emph{Ant}. iii) Transfer-learning tasks: A simple mobile robot with goal switching, and backward movement on \emph{Ant} and \emph{Snake}.

\paragraph{Algorithms:}
We compare RoMBRL with ME-TRPO, TRPO, and PPO. We acknowledge that training a recurrent policy network might be computationally expensive and susceptible to local optima. Therefore we also propose a variant of RoBMRL which uses a multilayer perceptron network (MLP) for policy representation. The MLP policy receives current observation as input and does not rely on history. Therefore this variant, named \emph{myopic RoBMRL} (BNN + MLP), can not handle \emph{planning under partial observability}. In particular, our standard RoMBRL uses Bayesian Neural Network (BNN) and LSTM (BNN + LSTM). All versions can also be accompanied with the online update ability as described in the previous section. 

\subsection{Fitting Toy Dynamics}

 \begin{wrapfigure}{r}{0.55\textwidth}
  \centering
  \includegraphics[width=0.55\textwidth]{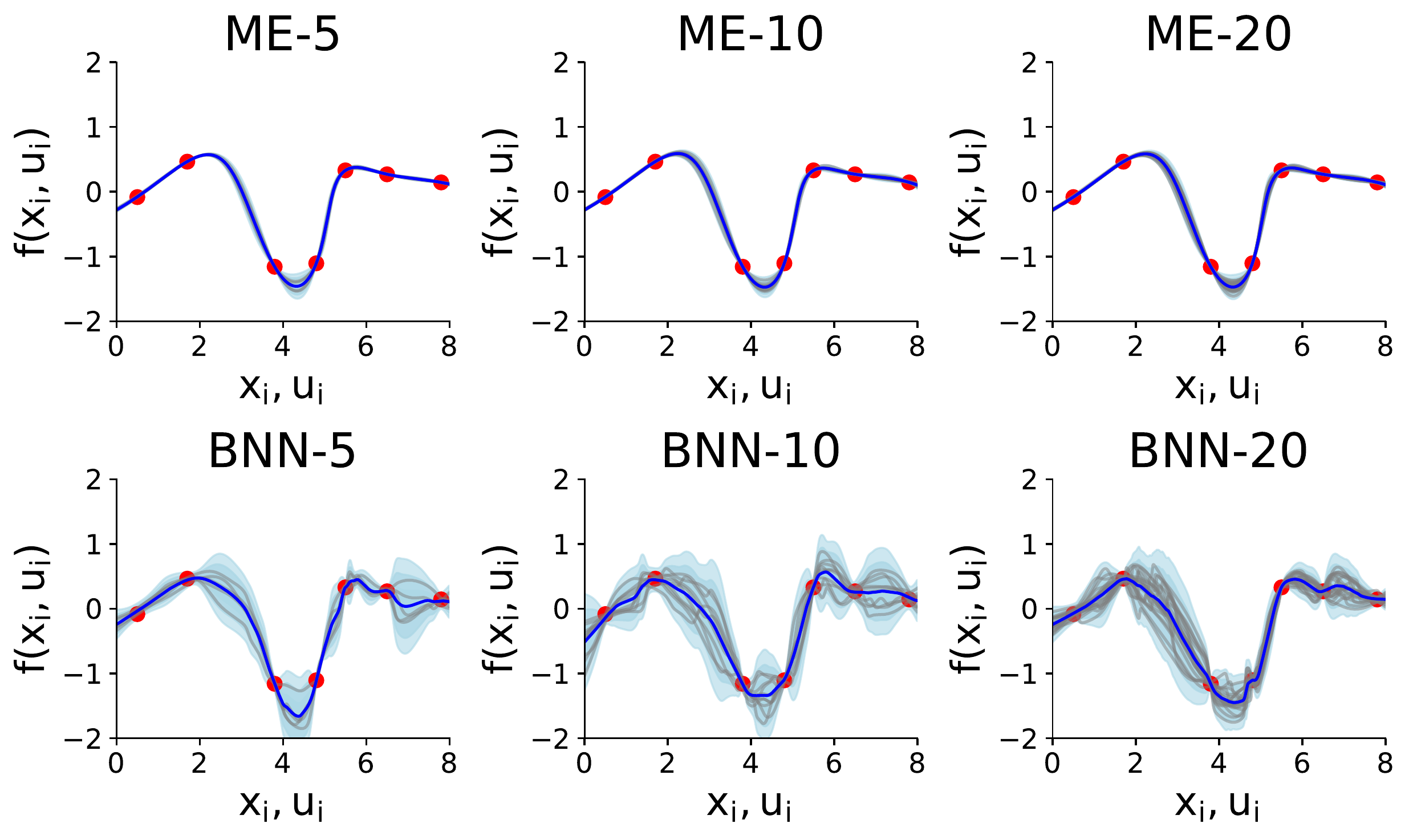}
  \caption{Evaluation on learning a toy dynamics with a varying number of models $k=5,10,20$.}
  \label{fig:exp1}
\end{wrapfigure}

This experiment is employed to evaluate i) the collapsing behaviour of the model-ensemble (ME) approach which is used by ME-TRPO and ii) how BNN overcomes the collapsing failure. Each kind of networks is evaluated under a varying number $k$ of models in the ensemble (on ME) or sampled models (on BNN). Each model in ME is represented with a feed-forward network of different initialised parameters. BNN samples are sampled via SGHMC.


As illustrated in Fig.~\ref{fig:exp1}, though being initialised differently and even increasing the number of models up to 20, all dynamics trained by ME are collapsed. In contrary, each BNN sample acts differently in an unknown region, thus clearly creates more meaningful samples which help obtain robust uncertainty estimation.

\subsection{Mujoco Control Tasks}

\begin{figure}   
  \centering
  \includegraphics[width=0.47\textwidth]{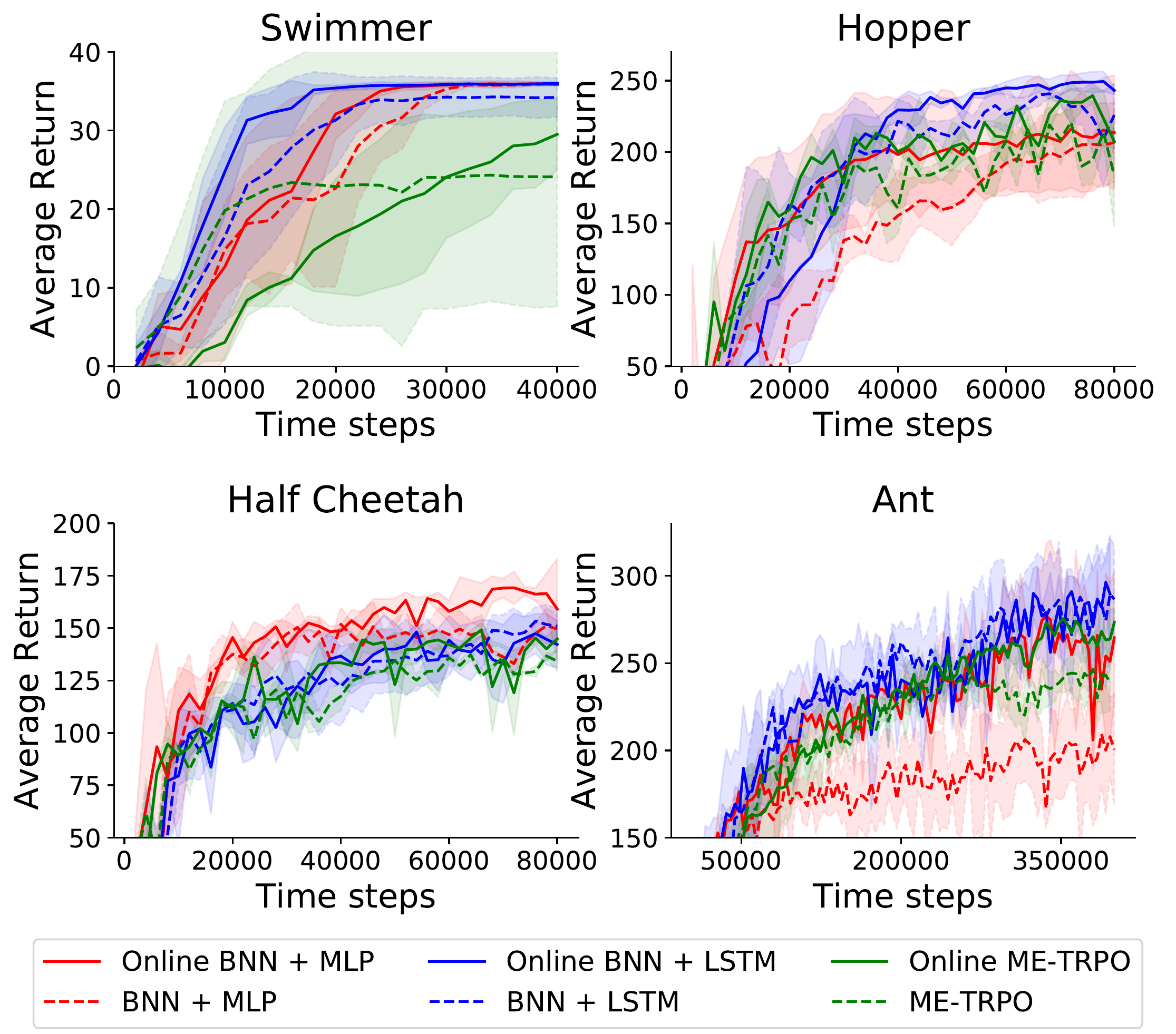} \includegraphics[width=0.47\textwidth]{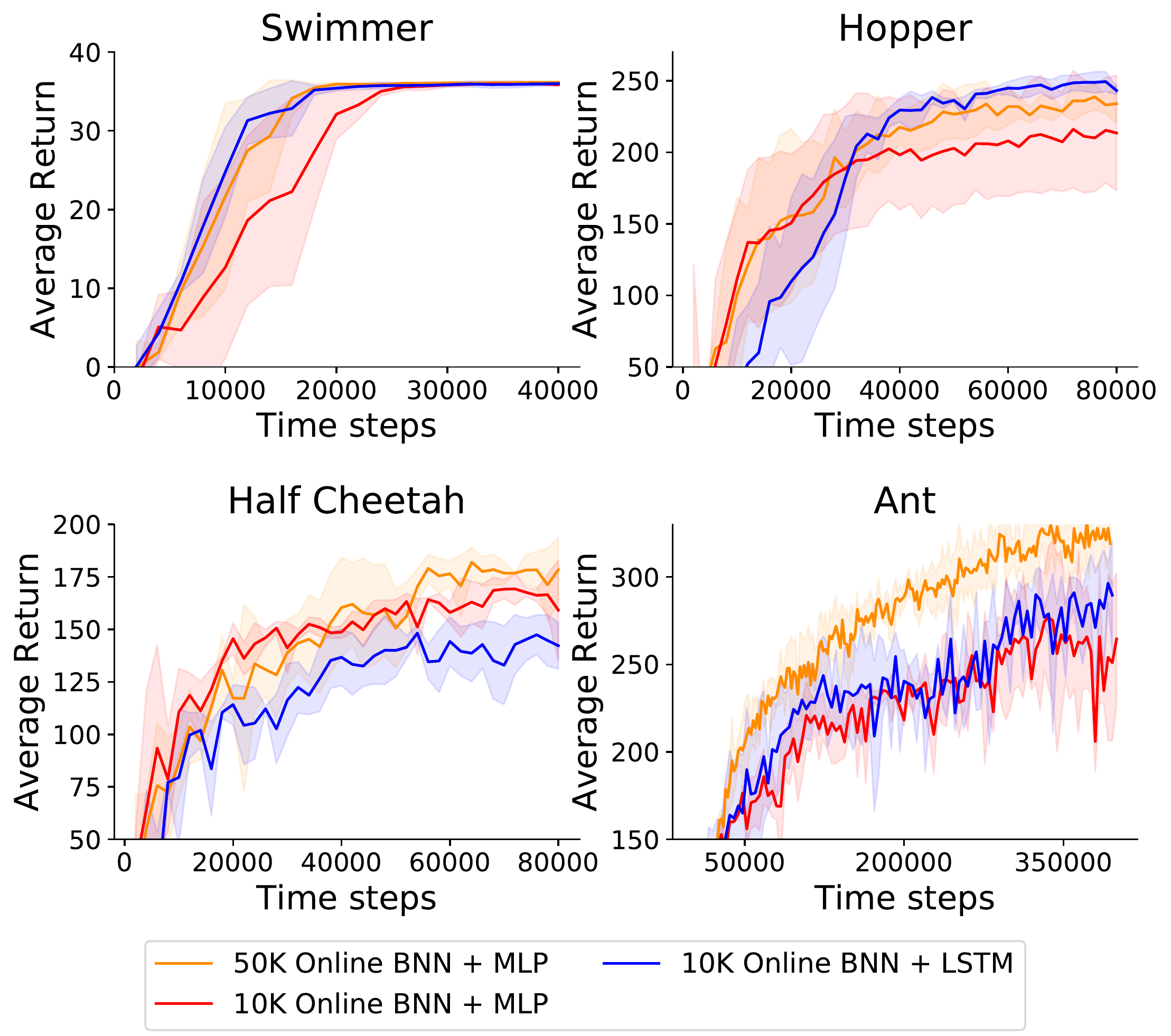}
  \caption{A full comparison between model-based approaches.}
  \label{fig:forward_fair}
\end{figure}

Figure~\ref{fig:forward_fair} (two columns on the left) show a comparison between model-based methods when they receive all similar settings regarding to the number of samples for training. The results have shown that our propoesd method (BNN+LSTM) outperforms myopic methods (BNN+MLP) and ME-TRPO. We have tried to increase the number of fictitious samples of the myopic methods (BNN+MLP) to 5 times, as depicted in the orange plots. As shown in Figure~\ref{fig:forward_fair} (two columns on the right), the performance of RoMBRL (BNN+LSTM) using 5 times fewer samples is still comparable. These results demonstrate that both high-quality uncertainty estimation (via Bayesian model learning) and forward-search with uncertainty propagation help find a better policy. 

\begin{wrapfigure}{r}{0.55\textwidth}
  \centering
  \includegraphics[width=0.55\textwidth]{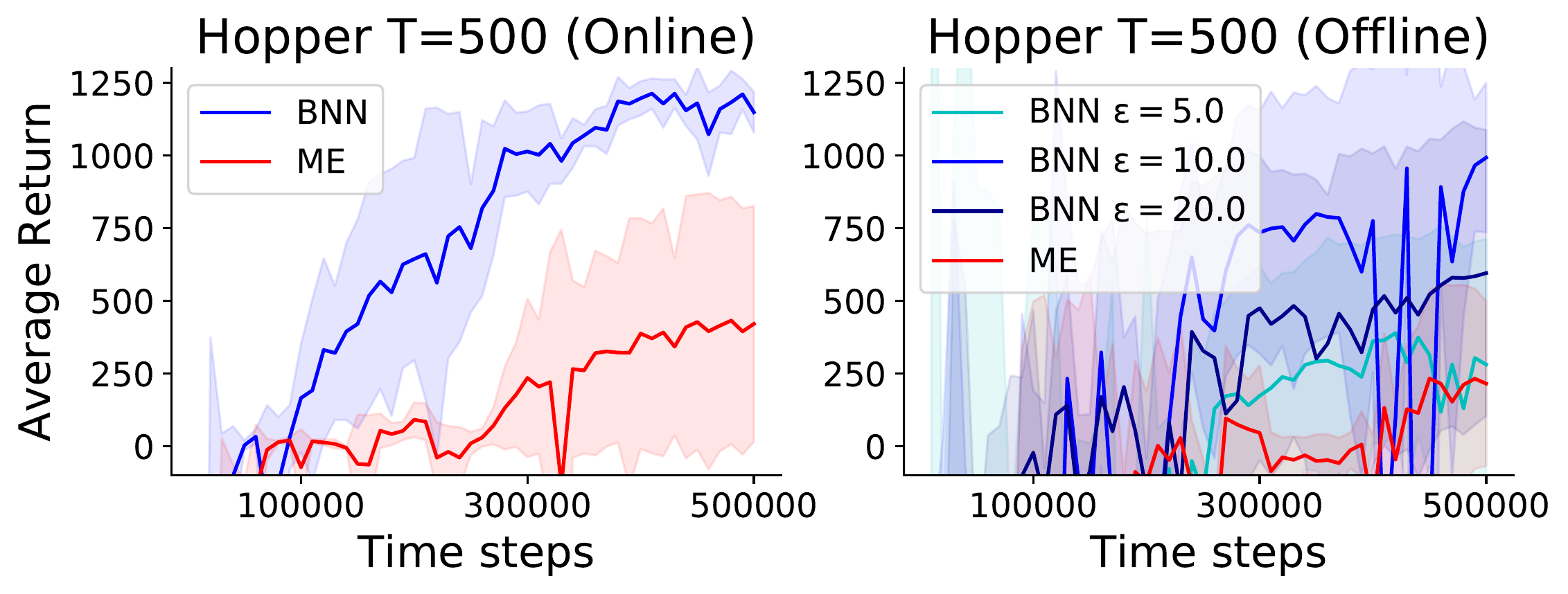}
  \caption{An ablation comparing RoMBRL(BNN) vs. ME-TRPO on a long-horizon Hopper with $T=500$.}
  \label{fig:hopper_long}
\end{wrapfigure}

We additionally evaluate how uncertainty would affect planning by creating a long-horizon Hopper task, $T=500$. Figure~\ref{fig:hopper_long} shows the comparison between our online version vs. ME-TRPO (left picture). This comparison shows how the inaccurate uncertainty estimation propagates over long planning horizon and affects the task performance. The right picture also shows how our Offline version is not stable to long-horizon planning given different settings of step size $\epsilon =5, 10, 20$ (controlling the burning time of SGHMC). The reason is due to non-robust uncertainty estimation when SGHMC is still struggling to converge. This result suggests that our online version can stabilise the model training, hence improve uncertainty estimation and leads to performance improvement.

\subsection{Transfer Learning Tasks}





\begin{figure}
  \centering
  \quad
  \includegraphics[width=0.6\textwidth]{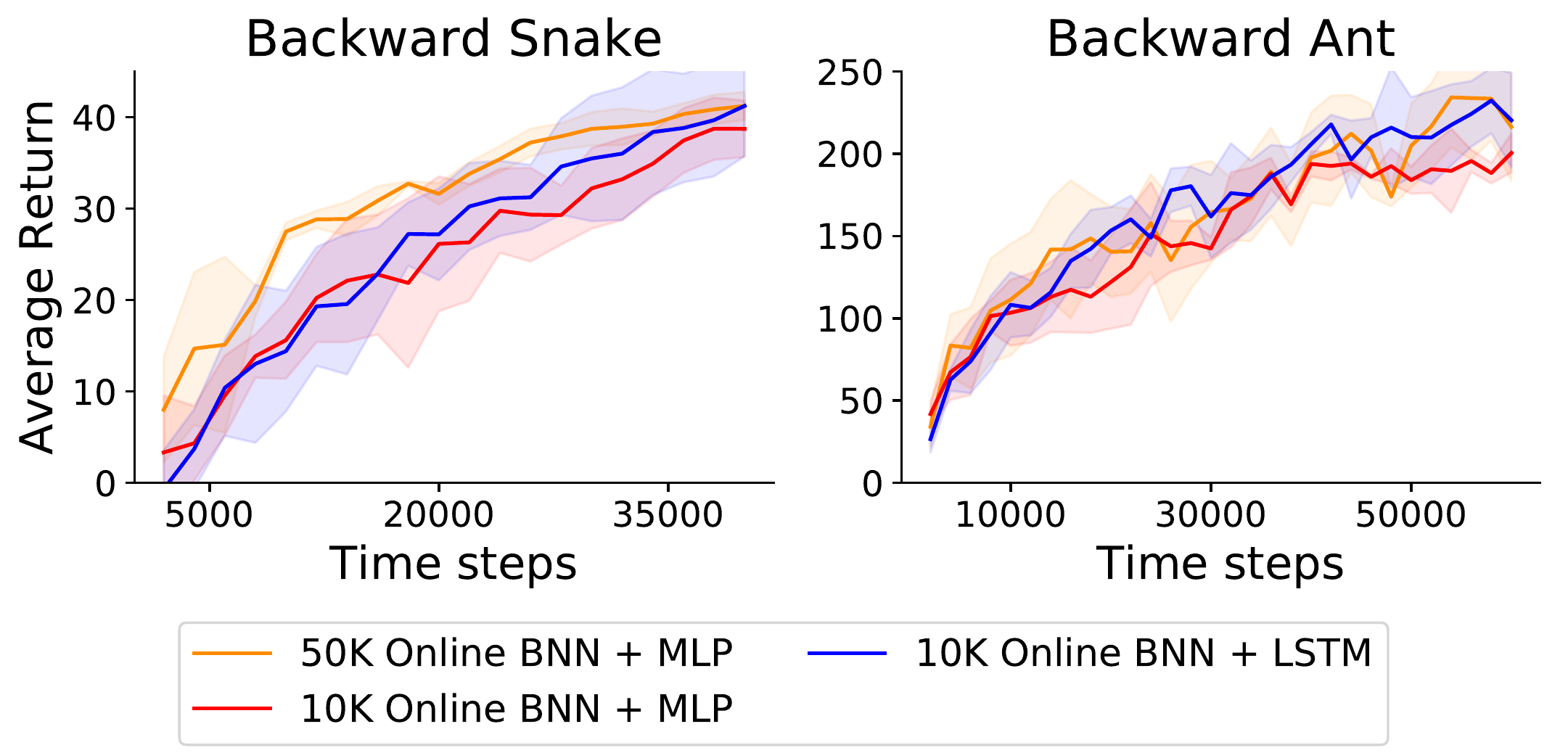}
  \caption{Comparisons with a different number of fictitious samples used for policy optimisation on two backward movement tasks.}
  \label{fig:backward_2}
\end{figure}

In this final task, we evaluate how robust uncertainty estimation helps improve task performance and enables transfer-learning more efficiently across tasks. The results shown here are for learning on the new task (new goal for the goal switching task; or go backward in the locomotion tasks) after the model distribution and the policy are trained till convergence on a previous task. Note that when the task is switched, by design both the proposed online model learning and robust uncertainty estimation (not collapsing) are able to help the policy and model quickly get updated with new data from the new task.
  \begin{figure}
  \centering
\includegraphics[width=0.9\textwidth]{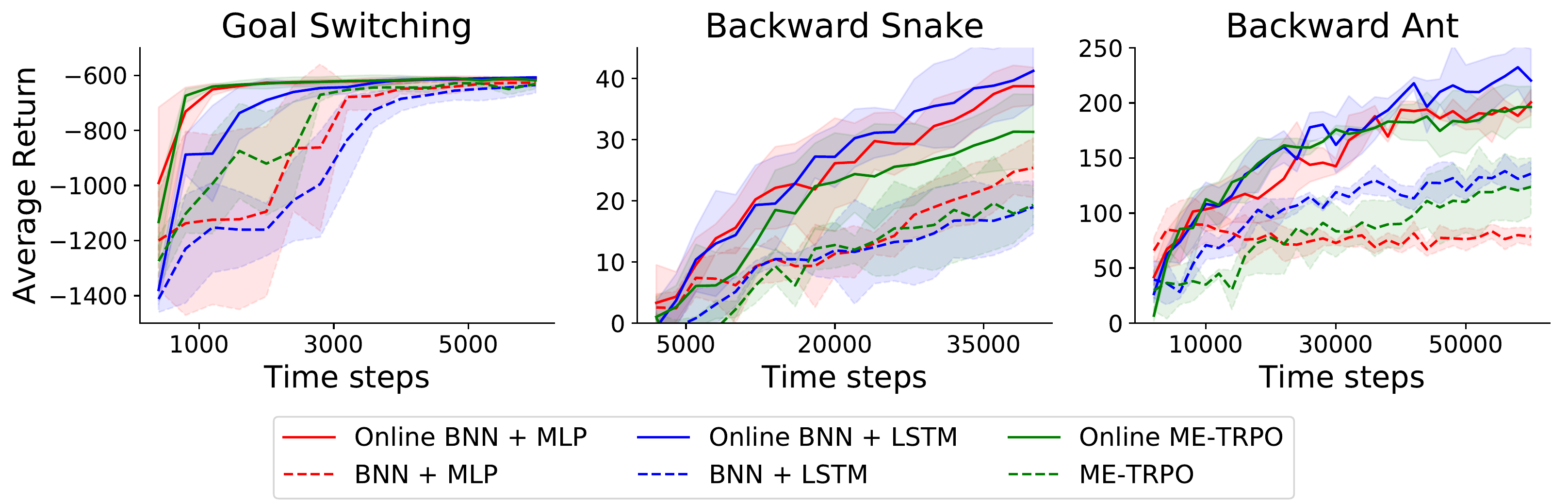}
  \caption{The goal switching and two backward movement tasks.}
  \label{fig:backward}
\end{figure}
Figure~\ref{fig:backward} depicts that the proposal of the online update rule helps significantly boost the performance of the standard RoMBRL. Interestingly, on a stable but complex RL task, Backward Snake, the online RoMBRL version converges slightly faster than the online ME-TRPO version. This result shows how the problem of model collapsing in ME methods affects task performance. Meanwhile, RoMBRL is still able to maintain uncertainty over unvisited state regions. Therefore RoMBRL can continue exploring these regions in the new task. In Ant, due to the instability of the task itself, exploration/exploitation trade-off becomes more unpredictable. Just little inaccurate prediction can lead the robot to an unstable state, which explains why the average return of BNN and ME are comparable. Finally, Figure~\ref{fig:backward_2} shows a comparison when we increase the number of samples in the myopic RoBMRL version to 5 times (50k vs. 10k). 


\section{Conclusions}
In this paper, we propose a principled Bayesian model-based reinforcement learning algorithm with deep neural networks, called \emph{RoMBRL}. Uncertainty estimation is the key to achieve an optimal trade-off between exploration/exploitation. We use a Bayesian neural network for model learning and simulation-based to propagate uncertainty. For scalability and flexibility, we represent action selection with a recurrent neural network, instead of a value function approximation. The results show that our proposed principled Bayesian BMRL can reason over uncertainty about the environment dynamics to reduce sample complexity when compared to recent methods of the same kind. A potential research direction could be improvements on how to scale RoMBRL to larger problems. In particular,  we can try to use more efficient online Bayesian learning techniques. In addition, a direction would be to improve the way how uncertainty can be propagated during planning.

\bibliography{my}

\section{Appendix}

\subsection{Environments}
We now describe three experiment sets.
\begin{enumerate}
\item {\bf Fitting Toy Dynamics}: We aim to analyse the behaviour of our Bayesian method on learning a simple transition distribution where only little data are observed. This domain is called \emph{1D Regression} where a data-set of tuples $\{x_i,y_i=f(x_i,u_i)\}$, where assuming $u_i$ is uncontrolled.
\item {\bf Mujoco Control Tasks}: They are Mujoco locomotion tasks on OpenAI Gym \cite{1606.01540}, \emph{Swimmer}, \emph{Hopper}, \emph{Half Cheetah} and \emph{Ant}. 
\item {\bf Transfer-Learning Tasks}: We do an evaluation on a transfer learning scenario in which we train ME-TRPO or RoMBRL to convergence on one task, then \emph{warm-start} them on a modified task (i.e. with the same underlying MDP dynamics but a different reward function). Our warm-start setting is to keep the model but re-learn the policy. In particular, we drop the policy learnt on the previous run (on the original task), while we keep the entire collected data and the learned dynamical models when learning on the modified task. We evaluate on three domains: 1) \emph{Differential Drive with Goal Switching} as depicted in Figure \ref{fig:diffdrive}; 2) forward movement learning on \emph{Snake} and \emph{Ant}, then switch to backward movement learning.
    
    \begin{figure}[H]     
      \centering
      \includegraphics[width=0.6\textwidth]{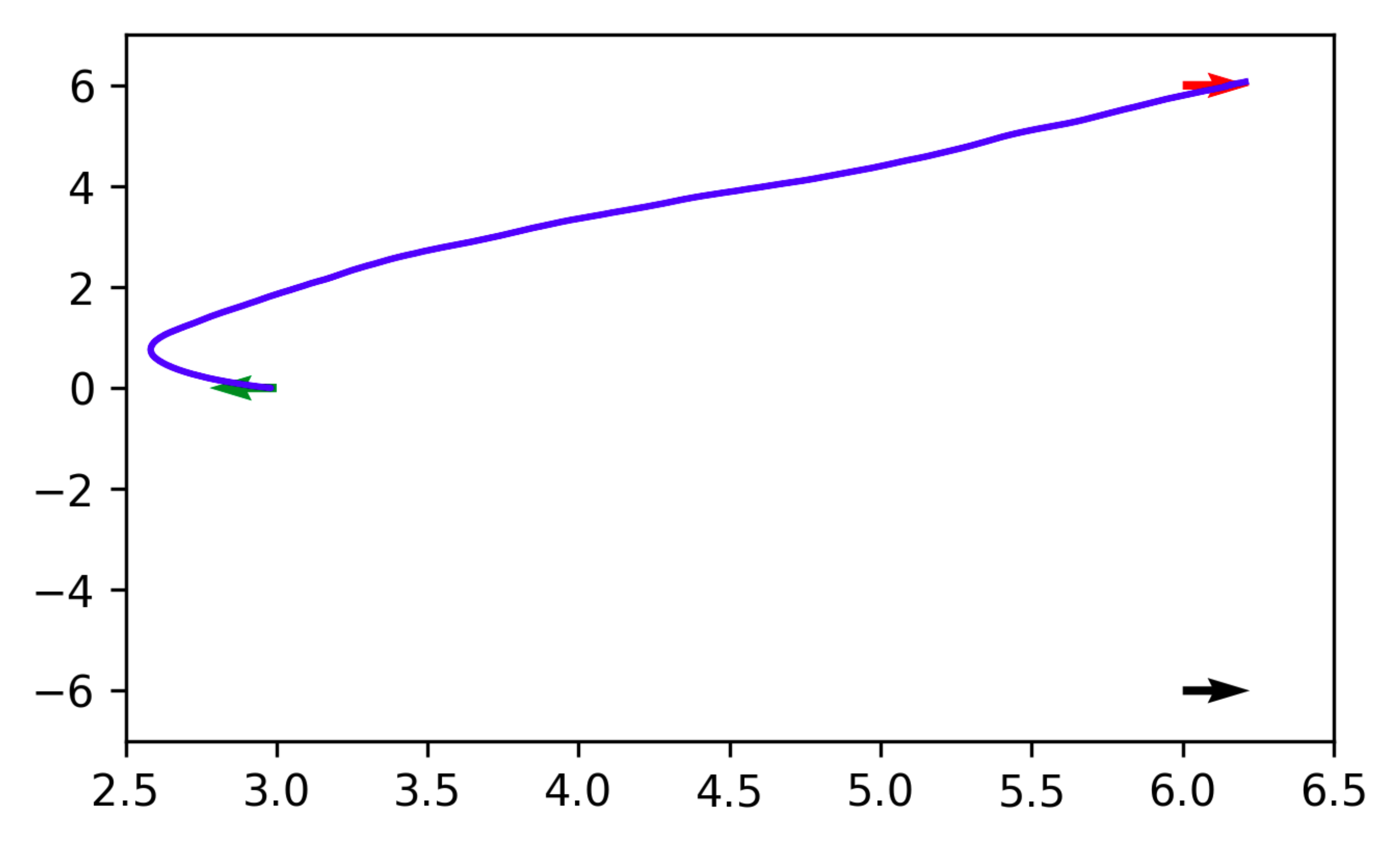}
      \caption{In the goal switching task, a \emph{differential drive robot} was first learned how to reach the original goal (black), then, it further explores the observation space to learn how to move from the original start position (green) to a newly defined goal (red). The blue line indicates the final propagated trajectory from the learned dynamic models.}
      \label{fig:diffdrive}
    \end{figure}
    
    \begin{itemize}
        \item Differential drive robot: This task has a simple non-linear dynamic that mimics a 2D navigation task of a non-holonomic mobile robot. The observations are $s=[p, \dot{p}]$, where $p = [x, y, \theta]$ respectively indicates the position and orientation of a mobile robot on the Cartesian space, while $\dot{p}$ represents the velocity. The action we can send to the robot is the linear and angular velocity $u=[v, \omega]$. Moreover, we adopt the go-to-goal reward function $r(s, u) = \|[x,y] - [x^g, y^g]\|^2 + 0.05\|u\|^2$ where a large penalty $-5000\|p - p^g\|^2$ is imposed at the end of the time horizon. Figure \ref{fig:diffdrive} illustrates the setup for this environment, where $^g$ denotes the state of the goal.        
        \item In backward locomotion domains on \emph{Snake} and \emph{Ant} on OpenAI Gym: We flip the speed observation of the reward function of the forward-walking tasks (default setting on OpenAI Gym) from positive to negative. 
    \end{itemize}
\end{enumerate}

\subsection{Experiment Setup}

The source code of this paper is publicly available on \url{https://github.com/thobotics/RoMBRL}. More specifically, we adopted some libraries in this implementation included {PYSGMCMC}\footnote{https://github.com/MFreidank/pysgmcmc} package for the implementation of Adaptive SGHMC method from Springenberg et. al. \cite{NIPS2016_6117}, \emph{rllab}\footnote{https://github.com/rll/rllab} for the evaluation on Mujoco domains and finally \cite{me-trpo} for the implementation of ME-TRPO baseline. 

\paragraph{Objective Functions}
To train the generative model, we employed in all experiments the potential energy function that is similar to the one proposed in Chen et. al. \cite{sghmc}, 

\begin{equation}
    \log p(\theta | s,a, s') \propto \log\big( p(s' = f_\theta(s,a) | s,a)p(\theta) \big)
\end{equation}

where we assume $f_\theta(s,a)$ follows a hierarchical Gaussian generative model with a Normal-Gamma distribution, and assuming a diagonal covariance matrix $Q$ (like in {PYSGMCMC} \cite{NIPS2016_6117}).

\begin{itemize}
    \item $p(y = f_\theta(s,a) | s,a) \sim \mathcal{N}(f_\theta(s,a), Q)$ \\
    \item $\theta \sim \mathcal{N}(0, \lambda)$ \\
    \item Gamma hyperprior: $Q^{-1} \sim \Gamma(a_1, b_1)$ \\
    \item Gamma hyperprior: $\lambda^{-1} \sim \Gamma(a_2, b_2)$
\end{itemize}


Finally, the negative log posterior that we used to simulate the Hamiltonian dynamics is:

\begin{equation}
\label{eq:loss}
\begin{aligned}
    \mathcal{L} &= (f_\theta(s,a) - s')^T Q^{-1} (f_\theta(s,a) - s') + \log p(\theta) + \log p(\lambda) + \log p(Q)
\end{aligned}
\end{equation}

\paragraph{Model-Ensemble Approaches}
Each neural network is independently trained under the same negative posterior in Equation \ref{eq:loss}, which means each neural network is now represented for a probabilistic network with Gaussian density $\mathcal{N}(f_\theta(x), Q)$. This adaptive ME version can simultaneously capture both the epistemic and aleatory uncertainty like \cite{DBLP:journals/corr/abs-1805-12114}.

We assume that ME-TRPO maintains an ensemble $k=5$ of neural network instances, except cases with an explicit mention. In contrast, we maintain a single BNN in RoMBRL, however we assume at each iteration we sample $k=20$ models from this BNN. This setting roughly gives a comparable and fair amount of computation budget (at each model update round) between the two algorithms: run time for training $k$ network models in TRPO vs. run time for sampling a set of $k$ models in RoMBRL.

\paragraph{SGHMC Setting}
Note that, in order to effectively stabilise SGHMC on vastly different scale data, we re-scale all the values of  input data to the same fixed range $[-10, 10]$ before training. The batch size equal 100 and the hyperprior values are also fixed on all domains $a_1 = 1.0, b_1 = 1.0, a_2 = 1.0, a_2 = 5.0$. In addition, to improve the sampling speed of SGHMC, we also adopt Adam (normal training) to optimize 3000 iterations before doing burn-in with a step-size of $2e-3$ on additional 3000 iterations and then collecting the samples at each 200 steps. Practically, we found this simple pre-train procedure is able to guide the SGHMC to find a better initialisation, thus helps stabilising SGHMC.

\begin{table*}[h]
    \caption{Training parameters on MBRL}
    \label{tb:dyn_fw_params}
    \centering
    \begin{tabular}{cccccccc}
    \toprule
    Domains & Dynamics & Policy & step-size $\epsilon$ & Rollout & Outer & Samples & Inner \\ 
    \midrule
    \multirow{ 3}{*}{\shortstack{Swimmer \\ (T=200)}} & BNN & MLP & 5.0 & 2000 & 20 & 5000 & 100 \\
    & BNN & LSTM & 5.0 & 2000 & 20 & 5000 & 100 \\
    & ME & MLP & NA & 2000 & 20 & 5000 & 100  \\ 
    \midrule
    \multirow{ 3}{*}{\shortstack{Hopper \\ (T=100)}} & BNN & MLP & 5.0 & 2000 & 40 & 15000 & 100 \\
    & BNN & LSTM & 5.0 & 2000 & 40 & 15000 & 100 \\
    & ME & MLP & NA & 2000 & 40 & 15000 & 100 \\ 
    \midrule
    \multirow{ 3}{*}{\shortstack{Half Cheetah \\ (T=100)}} & BNN & MLP & 5.0 & 2000 & 40 & 10000 & 100 \\
    & BNN & LSTM & 5.0 & 2000 & 40 & 10000 & 100 \\
    & ME & MLP & NA & 2000 & 40 & 10000 & 100  \\ 
    \midrule
    \multirow{ 3}{*}{\shortstack{Ant \\ (T=100)}} & BNN & MLP & 25.0 & 4000 & 100 & 10000 & 100 \\
    & BNN & LSTM & 25.0 & 4000 & 100 & 10000 & 100 \\
    & ME & MLP & NA & 4000 & 100 & 10000 & 100  \\ 
    \bottomrule
    \end{tabular}
\end{table*}

\begin{table*}[h]
    \caption{Training parameters on Online MBRL}
    \label{tb:dyn_bw_params}
    \centering
    \begin{tabular}{cccccccc}
    \toprule
    Domains & Dynamics & Policy & step-size $\epsilon$ & Rollout & Outer & Samples & Inner \\ 
    \midrule
    \multirow{ 3}{*}{\shortstack{Goal Switching \\ (T=200)}} & BNN & MLP & 1.0 & 400 & 15 & 4000 & 50 \\
    & BNN & LSTM & 1.0 & 400 & 15 & 4000 & 50 \\
    & ME & MLP & NA & 400 & 15 & 4000 & 50  \\ 
    \midrule
    \multirow{ 3}{*}{\shortstack{Backward Snake \\ (T=200)}} & BNN & MLP & 5.0 & 2000 & 20 & 10000 & 100 \\
    & BNN & LSTM & 5.0 & 2000 & 20 & 10000 & 100 \\
    & ME & MLP & NA & 2000 & 20 & 10000 & 100 \\ 
    \midrule
    \multirow{ 3}{*}{\shortstack{Backward Ant \\ (T=100)}} & BNN & MLP & 5.0 & 2000 & 30 & 10000 & 100 \\
    & BNN & LSTM & 5.0 & 2000 & 30 & 10000 & 100 \\
    & ME & MLP & NA & 2000 & 30 & 10000 & 100  \\ 
    \bottomrule
    \end{tabular}
\end{table*}

\subsection{Algorithm Setting}
\paragraph{Dynamic Model}

\begin{itemize}
    \item \emph{1D Regression}: In this task, we employ a simple MLP network with 2 FC layers, each has 50 units and use $tanh$ as activation function.     
    \item \emph{Mujoco Tasks}: Due to the model complexity, we change the activation to rectified linear unit (ReLU) and also increasing the number of units to 1024.
    \item \emph{Differential Drive Robot}: The network structure are the same as the one in the 1D Regression experiment except that the number of units has been increased to 100. 
\end{itemize}



\paragraph{Policy Model}

For simplicity, we employ the default settings of policy networks on all reinforcement learning domains similar to those of ME-TRPO \cite{me-trpo}, where the stochastic policy network is a simple Gaussian two layers feed-forward neural network (32 hidden units each and use $tanh$ as activation). In addition, the stopping condition of ME-TRPO was also adopted, in which we stop each model's policy training \cite{SchulmanLAJM15} when its reward does not increase for 25 inner iterations.


The details of the parameters settings on each domain are listed in Table \ref{tb:dyn_fw_params} and \ref{tb:dyn_bw_params}. In particular, \emph{step-size $\epsilon$} is the important hyper-parameter of SGHMC for training BNN, the lower the value the higher the exploration rate is. \emph{Rollout} is the number of datapoints executed on the real environment. Meanwhile, \emph{Samples} indicates the number of fictitious samples generated while doing policy optimization through TRPO. Finally, \emph{Outer} represents the number of iterations of the \emph{repeat} loop in Step 4 in Algorithm 2 (the overall number of model and policy gets updated. \emph{Inner} represents the iterations of the number of policy update via TRPO in Step 12.

In order to increase the training speed on domain Ant, we increase the training size to 4000 and use the step-size $\epsilon=$ 25.0 instead of 5.0. 



\subsection{Full results}
\subsubsection{Fitting toy dynamics}

\begin{figure}[H] 
  \centering
  \includegraphics[width=0.7\textwidth]{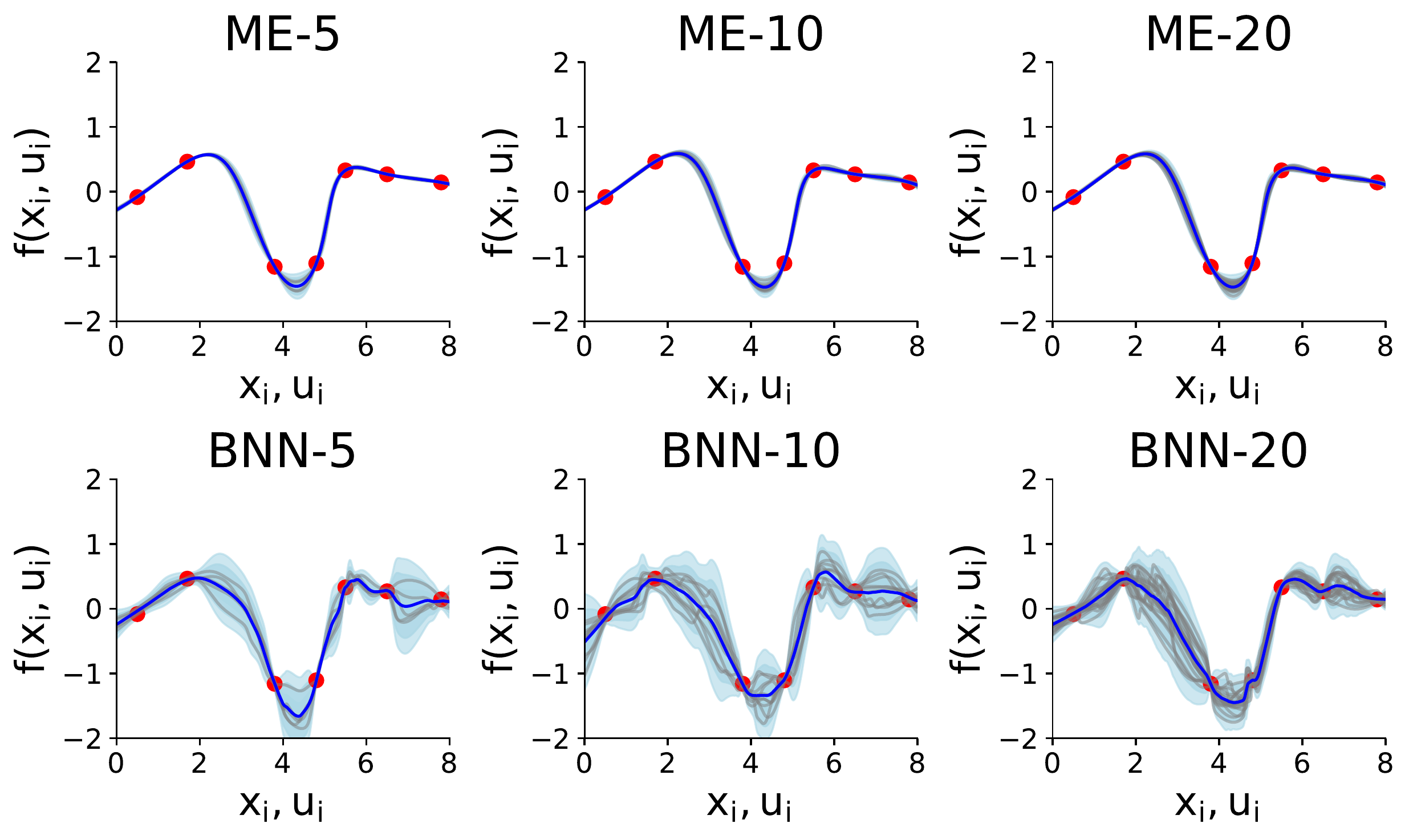}
  \caption{Evaluation on learning a toy dynamics with a varying number of models $k=5,10,20$.}
  \label{fig:exp1}
\end{figure}

In this experiment, the target is to learn a simple 1D transition function where only a small number of observations are given. This domain is employed to evaluate i) the collapsing behaviour of the model-ensemble approach and ii) how Bayesian Neural Network (BNN) overcomes this failure. Each kind of networks are being evaluated under a varying number $k$ of models (on ME) or sampled models (on BNN). Each model in ME is represented as one network with different initialised parameters. BNN samples are sampled via SGHMC.

As illustrated in Fig.~\ref{fig:exp1}, though being initialised differently and even increasing the number of models up to 20, all of the dynamics trained by the model-ensemble procedure are collapsed. In contrary, each BNN sample acts differently in an unknown region, thus clearly creates a more meaningful sample which leads to more robust uncertainty estimation.

\subsubsection{Mujoco control tasks}
We compare RoMBRL with ME-TRPO and two model-free methods (TRPO \cite{SchulmanLAJM15} and PPO \cite{SchulmanWDRK17}) on four Mujoco domains: Swimmer, Hopper, Half Cheetah and Ant.

\begin{figure}[H] 
  \centering
  \includegraphics[width=\textwidth]{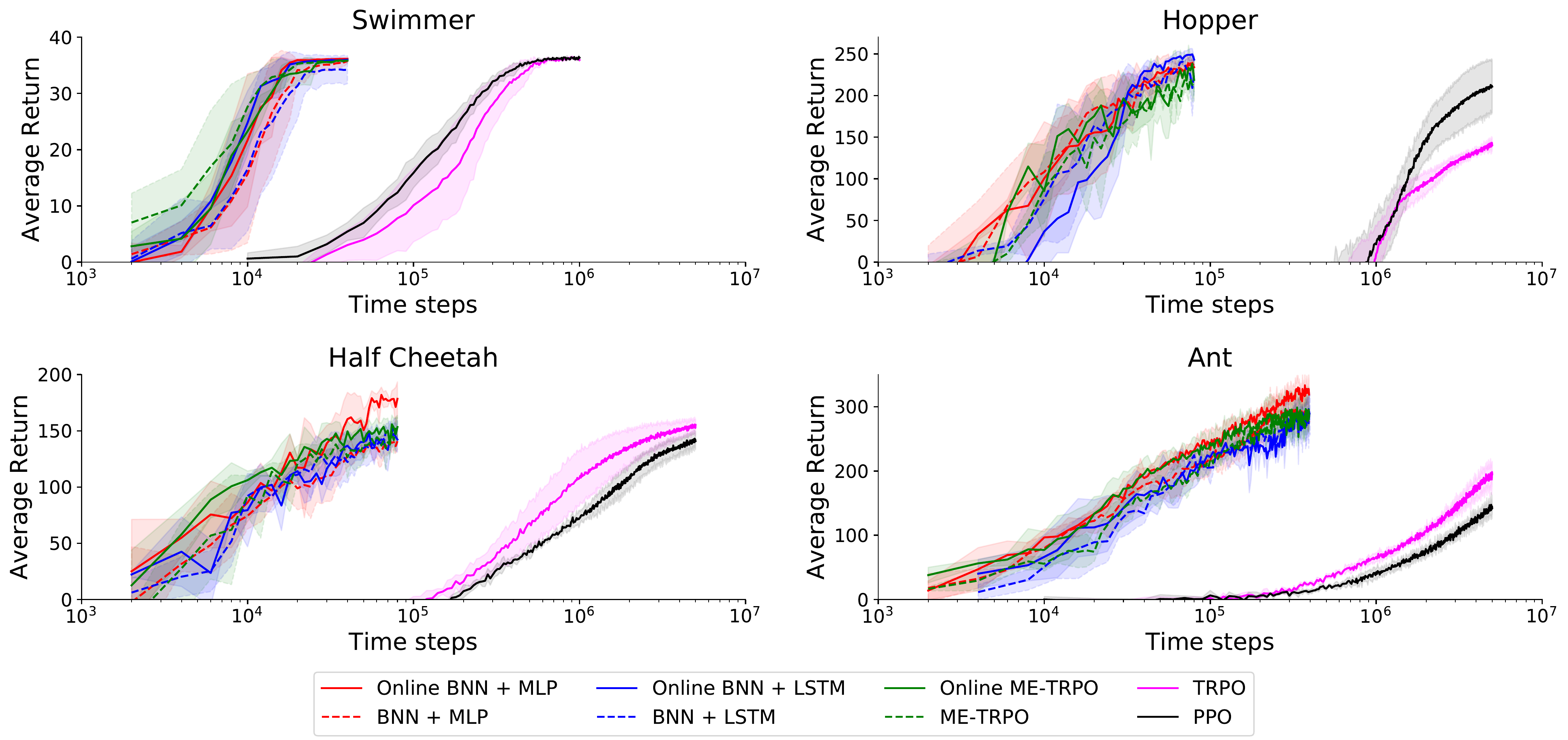}
  \caption{A full comparison between model-based and model-free approaches. Note that we use 5 times more fictitious samples to train all version without using LSTM.}
  \label{fig:forward}
\end{figure}

\begin{figure}[H] 
  \centering
  \includegraphics[width=0.6\textwidth]{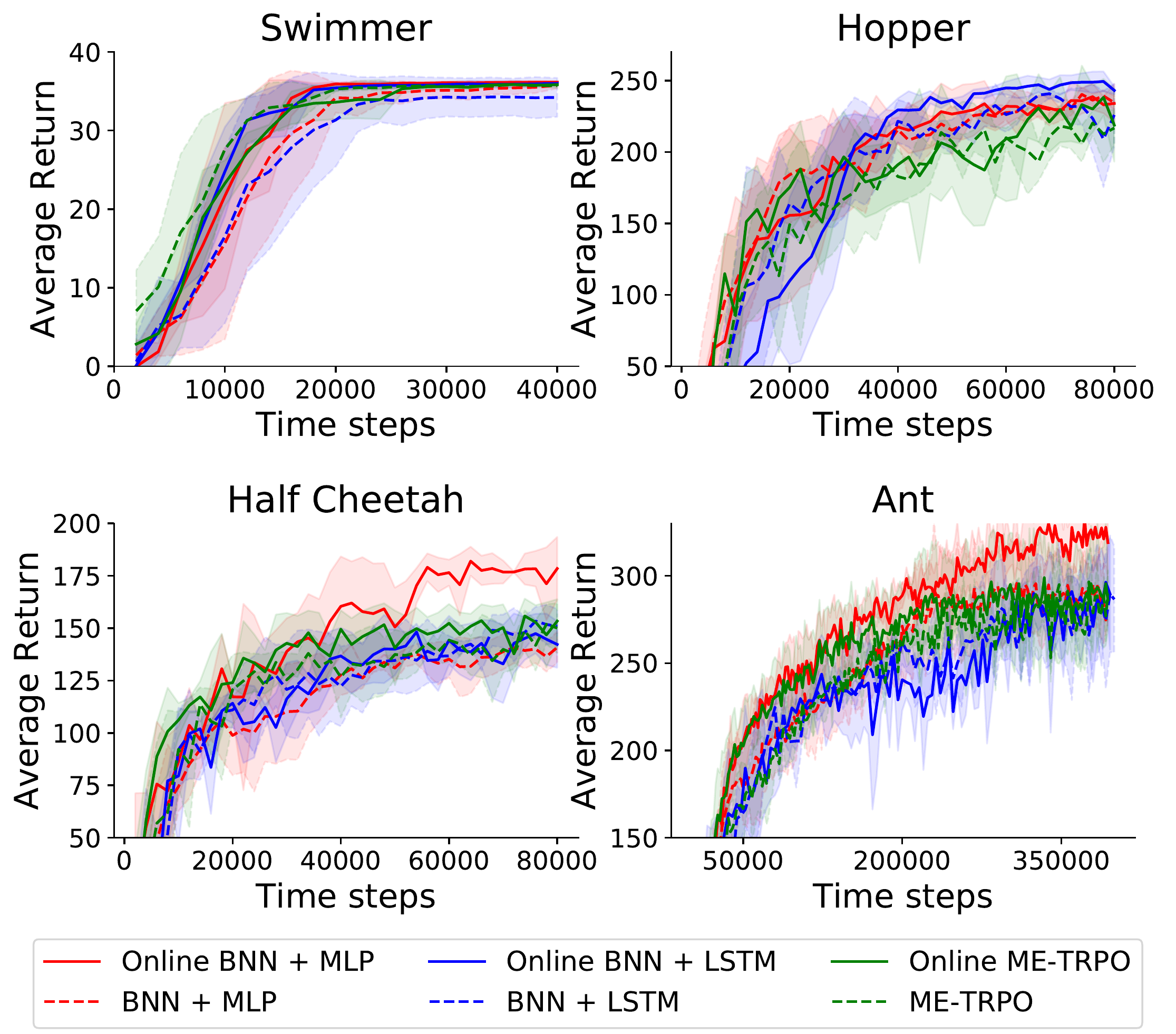}
  \caption{A full comparison between model-based. Note that we use 5 times more fictitious samples to train all version without using LSTM. Remove all model-free approaches.}
  \label{fig:forward2}
\end{figure}

\begin{figure}[H]  
  \centering
  \includegraphics[width=0.6\textwidth]{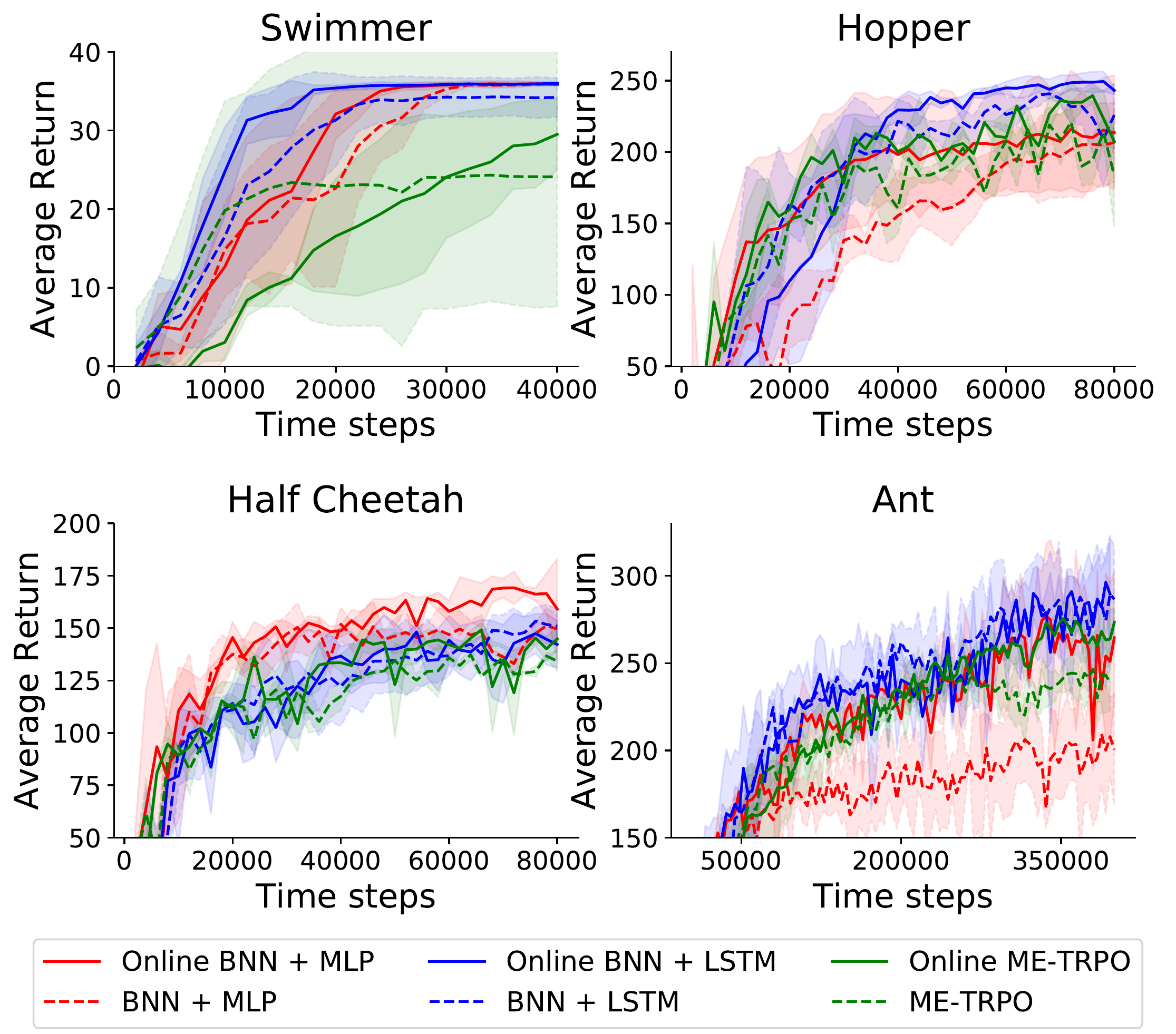}
  \caption{A full comparison between model-based. All methods use the same number of fictitious samples for policy optimisation.}
  \label{fig:forward_fair}
\end{figure}

\begin{figure}[H]
  \centering
  \includegraphics[width=0.6\textwidth]{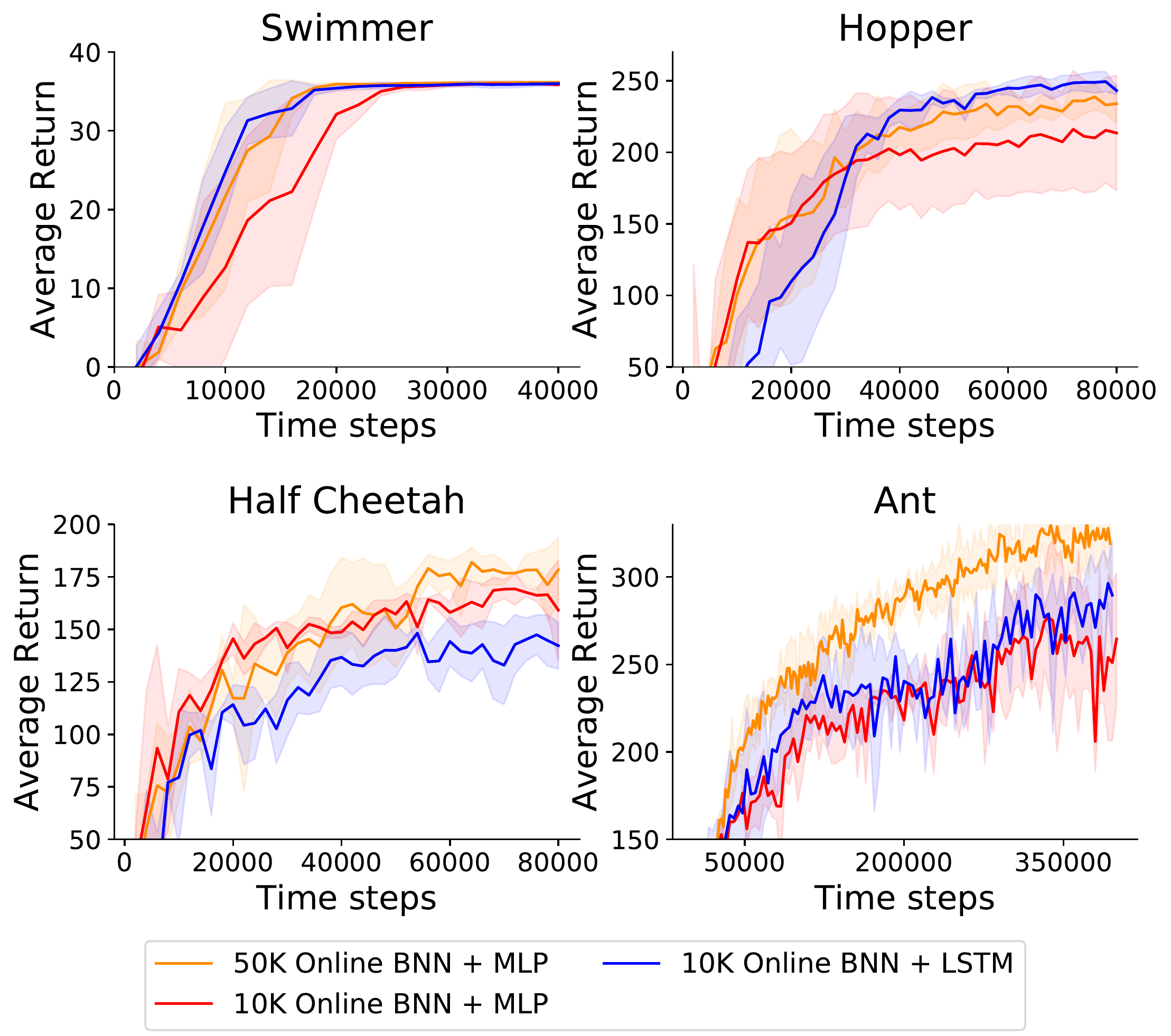}
  \caption{An ablation comparison between online Bayesian approaches.}
   \label{fig:forward_ablation}
\end{figure}

All model-based approaches substantially perform better than model-free approaches in these domains. Fig.~\ref{fig:forward} and Fig.~\ref{fig:forward2} depicts that the online version of BNN and ME perform slightly better than their standard counterparts in almost any Mujoco domains.

Figure~\ref{fig:forward_fair} shows a fair comparison between model-based methods when they receive all similar settings regarding to the number of samples for training. The results shown that our principled method (BNN+LSTM) outperforms myopic methods (BNN+MLP) and non-principled methods (ME-TRPO).

We have tried to increase the number of fictitious samples of the myopic methods (BNN+MLP) to 5 times. As shown in Fig.~\ref{fig:forward_ablation}, the performance of our principled RoMBRL (BNN+LSTM) using 5 times less samples is still comparable. These results demonstrate that both robust uncertainty estimation (via Bayesian model learning) and forward-search with uncertainty propagation help find a better policy.

\subsubsection{Transfer learning tasks}

\begin{figure}[H]
  \centering
  \includegraphics[width=\textwidth]{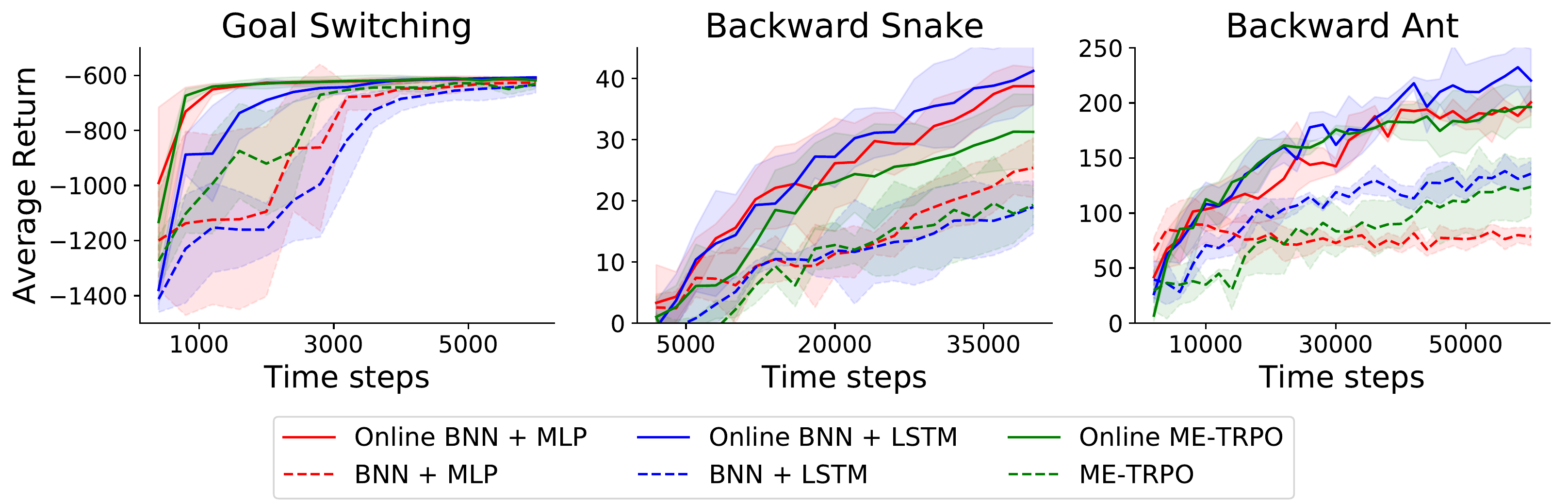}
  \caption{The goal switching and two backward movement tasks.}
   \label{fig:backward}
\end{figure}

\begin{figure}[H] 
  \centering
  \includegraphics[width=0.65\textwidth]{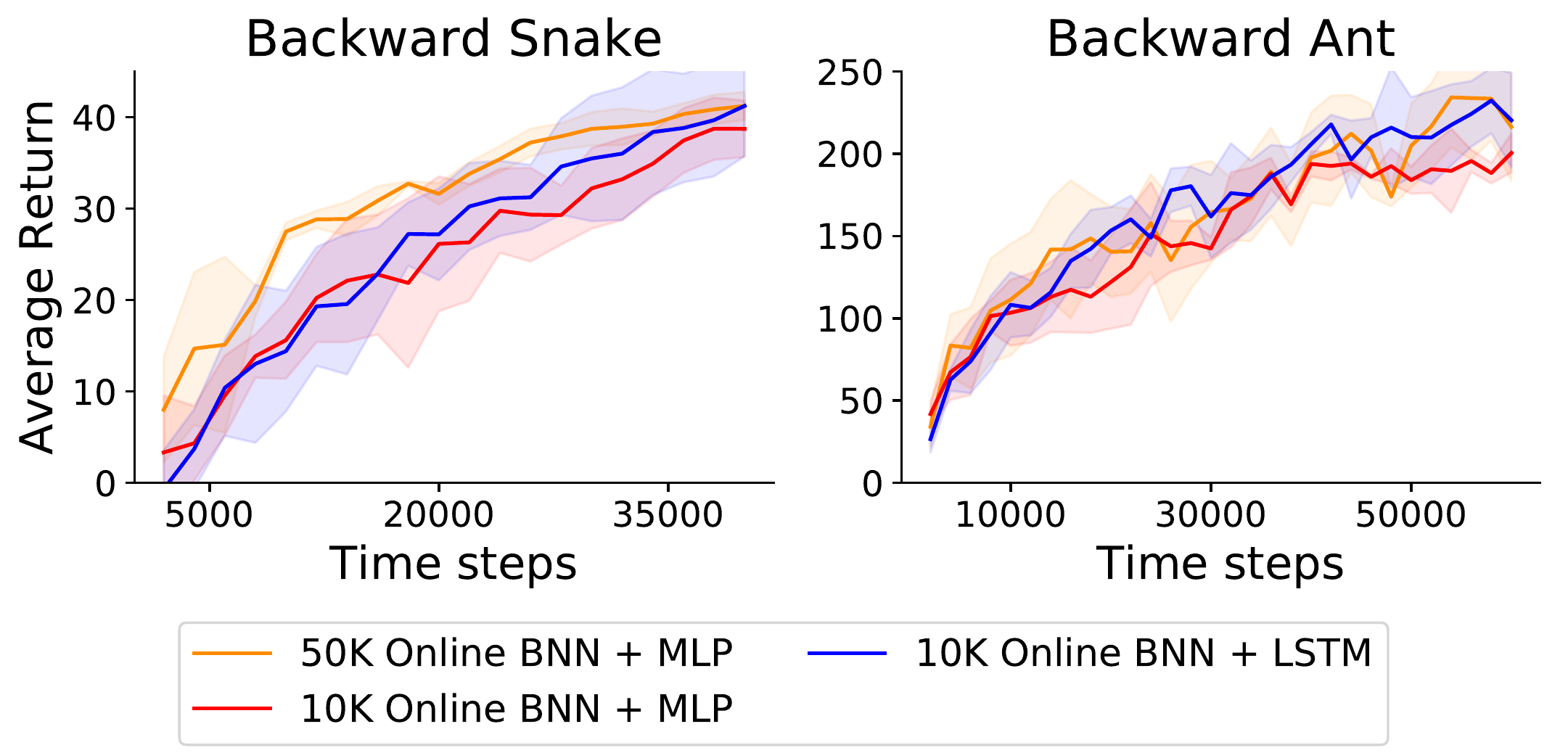}
  \caption{Comparisons with a different number of samples used for policy optimisation on two backward movement tasks.}
  \label{fig:backward_ablation}
\end{figure}

In this final task, we evaluate how Bayesian model-based RL can offer a robust model estimation that not only helps improve the task performance but also allows transfer-learning across tasks. The results shown here are for learning on the new task (new goal for the goal switching task; or go backward in the locomotion tasks) after the model distribution and the policy are trained till convergence on a previous task.

Figure~\ref{fig:backward} clearly depicts that the proposal of the online update rule helps significantly boost the performance of the standard RoMBRL. Especially, on a more challenging RL tasks like Backward Snake and Ant, the online RoMBRL version converges considerably faster than the online ME-TRPO version. This shows how the problem of model collapsing in ME methods affects the performance. Meanwhile, RoMBRL can still maintain robust uncertainty over unvisited state regions, therefore it is able to continue exploring these regions in the new task. 

Figure~\ref{fig:backward_ablation} show a comparison when we increase the number of samples in the myopic RoBMRL version to 5 times (50k vs. 10k). The result shows that the performance of the principled online RoMBRL (BNN+LSTM) is still very competitive.

\end{document}